\documentclass{article}
\usepackage{spconf,amsmath,graphicx}

\usepackage[rgb]{xcolor}
\usepackage{hyperref}
\hypersetup{
    colorlinks=true,
    linkcolor=blue,
    filecolor=magenta,
    urlcolor=cyan,
    citecolor=blue,
}
\usepackage{colortbl}

\usepackage{cite}
\usepackage{amsmath}
\usepackage{url}
\usepackage{amsfonts}
\usepackage{multirow}
\usepackage{makecell}
\usepackage{booktabs} % toprule, cmidrule, ...

\usepackage{pifont}
\newcommand{\cmark}{\ding{51}}
\newcommand{\xmark}{\ding{55}}

\usepackage{subcaption}

\newcommand{\fairseq}{\textsc{fairseq}}
\newcommand{\fairseqss}{\textsc{fairseq S$^2$}}

\usepackage{etoolbox}
\newtoggle{arxiv}
\toggletrue{arxiv} % uncomment this line to see the full version!

\title{A Holistic Cascade System, benchmark, and Human Evaluation Protocol for Expressive Speech-to-Speech Translation}

\name{\begin{tabular}{c}
    \it Wen-Chin Huang$^1$$^{\dagger\ddagger}$\thanks{$^{\dagger}$Work done while interning at Meta AI.~$^{\ddagger}$Equal contribution.}, Benjamin Peloquin$^2$$^{\ddagger}$, Justine Kao$^2$, Changhan Wang$^2$, \\
    \it Hongyu Gong$^2$, Elizabeth Salesky$^3$$^{\dagger}$, Yossi Adi$^2$, Ann Lee$^2$, Peng-Jen Chen$^2$
\end{tabular}}

\address{$^{1}$Nagoya University \\
  $^{2}$Meta AI\\
  $^{3}$Johns Hopkins University}

\begin{document}
\ninept
\maketitle

\begin{abstract}
Expressive speech-to-speech translation (S2ST) aims to transfer prosodic attributes of source speech to target speech while maintaining translation accuracy. Existing research in expressive S2ST is limited, typically focusing on a single expressivity aspect at a time. Likewise, this research area lacks standard evaluation protocols and well-curated benchmark datasets. In this work, we propose a holistic cascade system for expressive S2ST, combining multiple prosody transfer techniques previously considered only in isolation. We curate a benchmark expressivity test set in the TV series domain and explored a second dataset in the audiobook domain. Finally, we present a human evaluation protocol to assess multiple expressive dimensions across speech pairs. Experimental results indicate that bi-lingual annotators can assess the quality of expressive preservation in S2ST systems, and the holistic modeling approach outperforms single-aspect systems. Audio samples can be accessed through our demo webpage: \url{https://facebookresearch.github.io/speech_translation/cascade_expressive_s2st}.
\end{abstract}

\vspace*{-0.3cm}
\section{Introduction}
\label{sec:intro}

Speech-to-speech translation (S2ST) technologies are uniquely positioned to reduce communication barriers between speakers of different languages \cite{first-s2st}. To date, cascade systems consisting of automatic speech recognition (ASR), machine translation (MT), and text-to-speech (TTS) systems remain dominant in high-resource language pair settings. By contrast, direct S2ST methods do not rely on text generation as an intermediate step, and have been gaining attention due to their low computational cost and the ability to translate unwritten languages \cite{translatatron, translatatron2, direct-s2st-discrete, textless-direct-s2st}. However, these studies often focus on accurate semantic translation, overlooking the para-linguistic information in speech, which plays a crucial rule in speech communication.

Expressive S2ST aims to transfer para-linguistic attributes in source speech, such as intonation, emphasis, and emotion, to the generated speech while accurately translating between languages. This is an under-explored research area in multiple aspects. First, most research on expressive S2ST focuses on only one aspect of expressivity to the exclusion of other potential factors. Second, due to the difficulty of data collection, there lacks a benchmark test set to evaluate expressive S2ST systems. Missing too are standard, publicly available training sets. Finally, a single mean opinion score (MOS) test is often employed for subjective evaluation \cite{p808}, without clearly defining which aspects to evaluate.

\begin{figure}[t]
	\centering
	\includegraphics[width=\columnwidth]{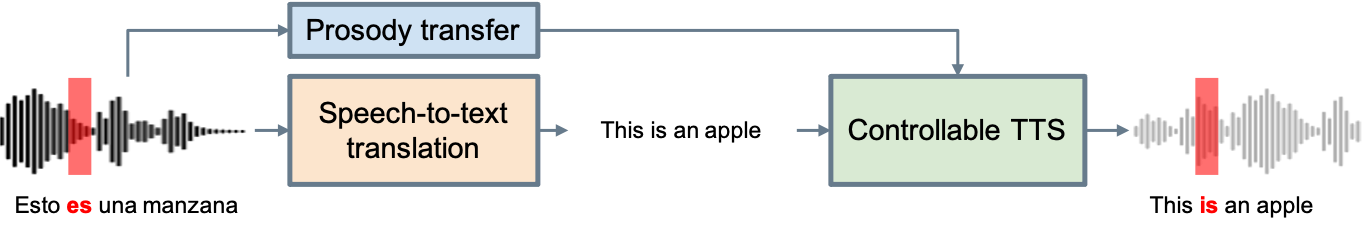}
	\caption{Illustration of the cascade expressive S2ST framework investigated in this work. It comprises an S2T model and a controllable TTS model. \label{fig:cascade-overview}}
	\vspace{-0.3cm}
\end{figure}

The goal of the current work is to accelerate research in expressive S2ST by addressing the above-mentioned limitations. Our contributions include:
\begin{itemize}
    \item A \textbf{holistic cascade system} for expressive S2ST. Specifically, our system consists an S2T model and a TTS model, as depicted in Fig.~\ref{fig:cascade-overview}. The core of our system is a controllable TTS that takes as input the translated text from a speech-to-text (S2T) model and multiple clues from the source speech. 
    \item A curated \textbf{benchmark} test set based on \textit{Heroes} \cite{heroes}, a dataset from an English TV series dubbed in Spanish.
    Our curation process includes, denoising, quality checks and transcription correction by human annotators.
    \item A novel \textbf{human evaluation protocol}, in which annotators are asked to provide pairwise similarity ratings across multiple expressive aspects including \textit{emphasis}, \textit{intonation}, \textit{rhythm}, and \textit{emotion}.
\end{itemize}

\vspace*{-0.3cm}
\section{Related works}

\subsection{Expressive S2ST}

Early approaches to expressive S2ST \cite{expressive-s2st-intonation} focused on transferring intonation. In this framework, source intonation attributes are automatically labeled, and then transferred to the target language using a statistical word alignment model. These attributes, along with the translated text, are fed into a TTS model. Related works focused on the transfer of word emphasis, adopting similar techniques, first employing word alignment information, \cite{emphasis-s2st}, and then exploring a joint sequence-to-sequence based emphasis and content translation approach \cite{emphasis-s2st-seq2seq}. More recently, \cite{face-dubbing} utilized speech synthesis markup language to transfer intonation. This family of approaches all focused on a single aspect of expressivity in isolation. The current work expands on these earlier studies by incorporating multiple expressivity aspects simultaneously.

\subsection{Style transfer and controllable TTS}

Controllable TTS refers to synthesizing speech given a text and certain attributes. When given reference speech, it is also known as style/prosody transfer in TTS. There are two mainstream approaches to this task. The earliest attempt is the global style transfer framework \cite{prosody-transfer}, which tries to encode and aggregate the global attributes of the reference speech into a single fixed-dimensional embedding with different bottlenecks \cite{gst, global-vae, gmvae-tacotron, stylespeech}. By contrast, fine-grained style transfer extract frame-level embeddings to capture local prosodic properties. However, due to the length mismatch between the frame-level prosodic embeddings and the text sequence, techniques like an attention mechanism \cite{prosody-control-second-attention} or forced alignments \cite{prosody-control-forced-alignment} are required.

One disadvantage of fine-grained style transfer is that it can only be applied to same-text (parallel) transfer, where the content in the reference speech and the text are identical. Global style transfer does not share this limitation and can be used in non-parallel scenarios. A novel contribution of this work is applying global prosody transfer in the S2ST scenario, whose cross-lingual setting is different from all previous works.

\begin{figure}[t]
	\centering
	\includegraphics[width=0.8\columnwidth]{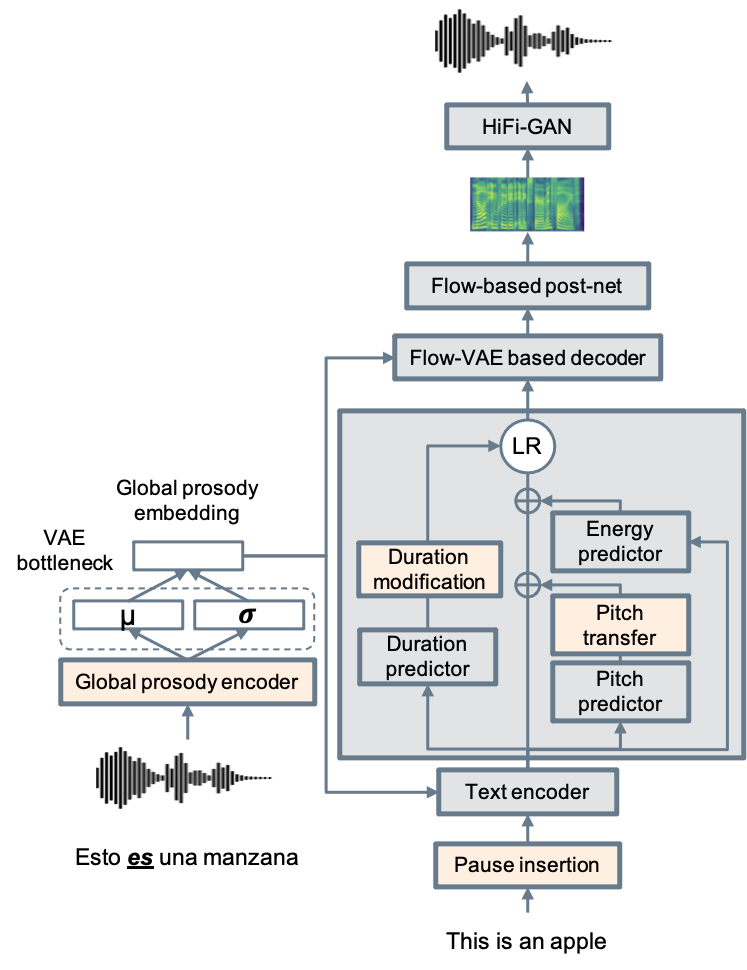} 
	\caption{Illustration of the proposed controllable TTS model, which is based on FastSpeech 2 \cite{fs2} and PortaSpeech \cite{ps}. Orange colored boxes indicate modules related to prosody transfer. \label{fig:model-overview}}
	\vspace{-0.3cm}
\end{figure}

\vspace*{-0.3cm}
\section{A holistic, cascade expressive S2ST system}
\label{sec:method}

As illustrated in Fig.~\ref{fig:cascade-overview}, we propose a cascade expressive S2ST system consisting of an S2T model and a controllable TTS model. In this work, we focus on designing the controllable TTS model to take multiple prosodic clues from the source speech. The controllable TTS model is illustrated in Fig.~\ref{fig:model-overview}. We directly employ an off-the-shelf S2T model, detailed in \cite{fairseq-s2t}.

\subsection{TTS model backbone}

Phoneme-based FastSpeech 2 \cite{fs2} is the main backbone for our controllable TTS model - it allows manual control of various factors, such as duration and pitch. To increase the model capacity and encourage prosodic diversity, we adopt a variational autoencoder (VAE) with an enhanced prior and a flow-based post-net proposed in PortaSpeech \cite{ps}. Training objectives include an L1 loss in the mel-spectrogram domain, an L2 loss for each of the duration, pitch and energy predictors, the KL loss for the VAE, and the negative log-likelihood for the flow-based post-net. At test, we use a pre-trained HiFi-GAN \cite{hifigan} for mel-spectrogram to waveform inversion.

\subsection{Global Prosody Transfer}

Following previous work, we first used a reference encoder to generate the global prosody embedding. The architecture design followed \cite{stylespeech} which consists of gated convolutional layers and self-attention layers. Following \cite{global-vae}, a VAE bottleneck was applied to improve the generalization ability. Instead of simply adding the embedding to the text encoder output, we used speaker adaptive layer normalization layers \cite{stylespeech} to better fuse the prosody embedding into the TTS model.

\subsection{Local Prosody Transfer}

\begin{figure*}[t]
	\centering
	
	\begin{subfigure}[b]{0.6\columnwidth}
		\centering
  		\includegraphics[height=2.5cm]{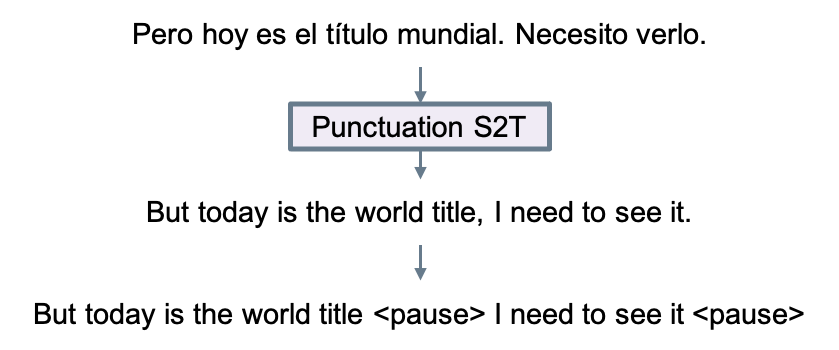}
		\caption{Pause insertion.}
   		\label{fig:pause-insertion}
	\end{subfigure}%
	\quad
	\begin{subfigure}[b]{0.7\columnwidth}
		\centering
	    \includegraphics[height=3.5cm]{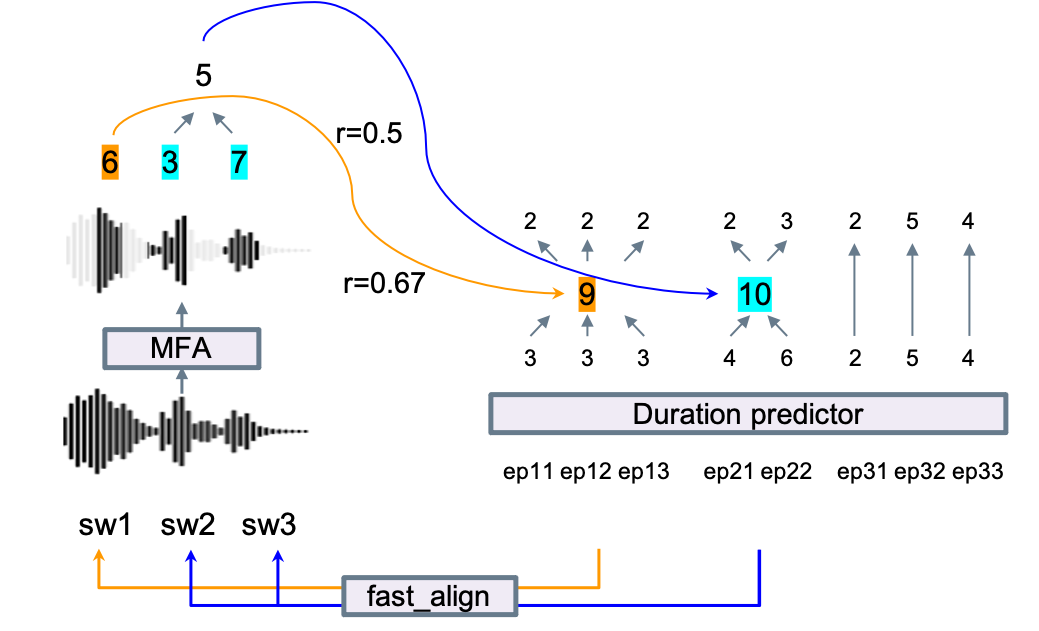}
		\caption{Duration modification.}
   		\label{fig:duration modification}
   	\end{subfigure}%
	\begin{subfigure}[b]{0.7\columnwidth}
		\centering
	    \includegraphics[height=3.5cm]{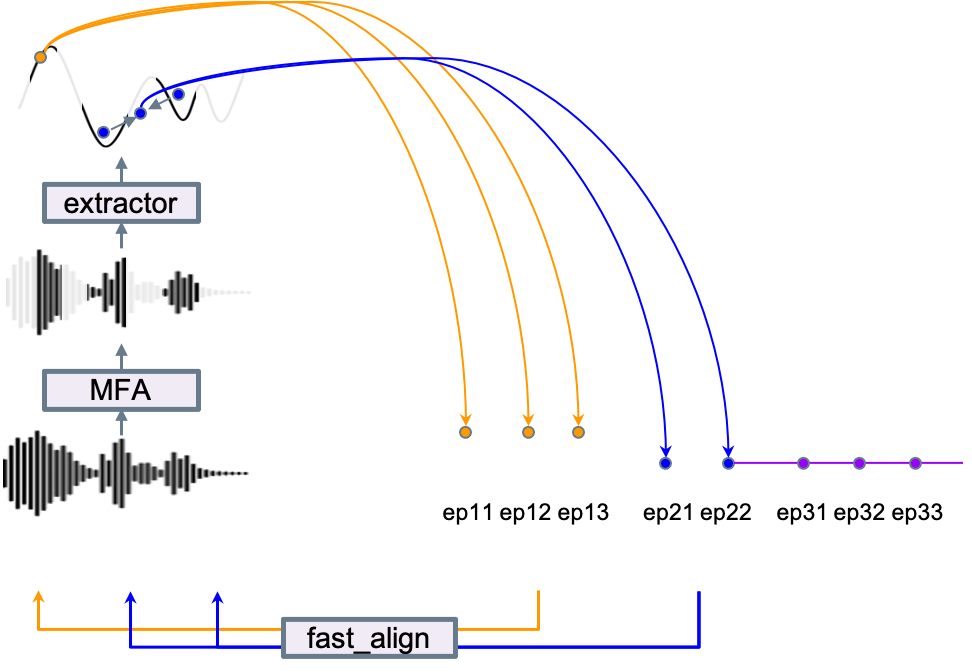}
		\caption{F0 transfer.}
   		\label{fig:f0-transfer}
   	\end{subfigure}\\
	
	\centering
	\caption{Illustration of the local prosody transfer techniques.}
	\vspace{-0.3cm}
	\label{fig:local-prosody-transfer}
\end{figure*}

In our initial experiments, we found that in the cascade S2ST framework, there are certain local attributes that the global prosody embedding fails to capture. We therefore propose a collection of local prosody transfer techniques, which is depicted in Fig.~\ref{fig:local-prosody-transfer}.  

% \subsubsection{Pause insertion}
\noindent{\textbf{Pause insertion.}}
Short pauses contribute to several factors in expressivity, including emphasis and rhythm. % Transfering pauses from the source language to the target language is a non-trivial task.
\iftoggle{arxiv} {
Our pause insertion technique, as depicted in Fig.~\ref{fig:pause-insertion}, is based on the observation that punctuations in written text often correspond to pauses in speech.
}{
Our pause insertion technique is based on the observation that punctuations in written text often correspond to pauses in speech.
}
We first train our S2T model with punctuation data (such that the output contains punctuations). Then, we simply convert the punctuations into pauses with a predefined duration (empirically set to 0.6 sec).

% \subsubsection{Duration modification}
\noindent{\textbf{Duration modification.}}
The duration of each spoken word (i.e. local speaking rate) contributes to emotion and rhythm.
\iftoggle{arxiv} {
Our proposed duration modification method is similar to \cite{expressive-s2st-intonation}, as depicted in Fig.~\ref{fig:duration modification}.
}{
Our proposed duration modification method is similar to \cite{expressive-s2st-intonation}.
}
First, we use a forced aligner to extract the duration of each word in the source speech. Then, an external MT word aligner was used to find word pairs between the source and target text, and calculate the ratios between the source word durations and the predicted word durations of the target words. Finally, the duration of each phoneme is modified according to the ratio. Note that this process is only applied to vowels.

% \subsubsection{Pitch transfer}
\noindent{\textbf{Pitch transfer.}}
Pitch contributes to emotion, emphasis and intonation.
\iftoggle{arxiv} {
Our pitch transfer method, as depicted in Fig.~\ref{fig:f0-transfer}, is similar to the duration modification technique.
}{
Our pitch transfer method is similar to the duration modification technique.
}
After extracting the word-level averaged f0, it is assigned to its corresponding word and phonemes. Note that for target language words without an MT alignment, we used bicubic interpolation to fill in the f0 values. We also applied a mean-variance speaker-level F0 normalization process.
% \al{I might have missed this, but have we mentioned speaker-level F0 normalization? (or did we end up not applying normalization?)}

\subsection{Implementation}
% \al{nit: clarify ``Spanish-English'' S2T model and ``English'' TTS model}
In our experiments, we focus on the Spanish-English (es-en) direction.
The es-en S2T model was trained on a series of open-sourced datasets\footnote{\label{s2t}Details can be found online: \url{https://huggingface.co/facebook/xm_transformer_600m-es_en-multi_domain}}.
The English TTS model was trained on the Blizzard Challenge 2013 segmented set \cite{bz2013}, which is a 10 hours single-speaker female audiobook dataset. We based our codebase on the \fairseq\ S2T and \fairseqss\ toolkits \cite{fairseq, fairseq-s2t, fairseq-s2}. The S2T model was open-sourced\footnote{See footnote \ref{s2t}.}. To train the TTS model, phoneme-level durations were extracted with a Montreal forced aligner (MFA) \cite{mfa} trained on VoxPopuli \cite{voxpopuli}. During test time, we used an in-house forced aligner for better performance. The pitch extractor was WORLD \cite{WORLD}, and word alignments was extracted using AWESOME \cite{awesome-align}.

\section{Datasets}

We curated a benchmark test set called the \textit{Heroes} dataset. Heroes is a sci-fi English TV series with Spanish dubbing \cite{heroes}. The English actor/actress and the Spanish dubber are always of the same gender. Since it contains background noise and music, we first applied DEMUCS, a denoising model \cite{demucs-se}. As the denoised audio can be flawed, we conducted human annotation to filter out samples with audio quality MOS score less than four and transcribed the samples.

We also explore a \textit{Mined Audiobook} dataset which exploits sentence embeddings in English and Spanish data from LibriVox \cite{LASER}\footnote{\url{https://github.com/facebookresearch/LASER/tree/main/tasks/librivox-s2s}}. Starting from the highest cosine similarity sample pairs, we manually picked sample pairs with the same gender. Since these samples are automatically mined, a human transcription process was also conducted.
Note that we do not consider the Mined Audiobook dataset as a benchmark test set, because the underlying prosodic alignment in the data is unknown and can be variable. Unlike the Heroes dataset which includes high-quality dubbing by actors throughout, we sampled the Mined Audiobook set by selecting samples with high pitch variability, then manually filtering to match gender and prosodic elements.
Table~\ref{tab:benchmarks} shows the summary of these two datasets.
\iftoggle{arxiv}{
More analyses on the Heroes dataset can be found in Appendix~\ref{sec:heroes_emotion_label}.
}
% \al{Is this from https://github.com/facebookresearch/LASER/tree/main/tasks/librivox-s2s? If yes, let's add a link in footnote.}
% \al{Are these samples from mined audiobook aligned in prosody?}

\begin{table}[t]
	\centering
	\caption{Summary of the datasets.}
	
	\centering
	\vspace*{-0.3cm}
	\begin{tabular}{ c | c c c c }
		\toprule
		\multirow{2}{*}[0pt]{Dataset} & \multirow{2}{*}[0pt]{Domain} & \multirow{2}{*}[0pt]{\# samples} & \multicolumn{2}{c}{Duration} \\
		& & & Mean & Median\\
		\midrule
		Heroes & TV series & 406 & 3.10 & 2.79 \\
		\makecell{Mined\\Audiobook} & Audiobook & 100 & 6.03 & 5.32 \\
		
		\bottomrule
	\end{tabular}
	\vspace*{-0.5cm}
	\label{tab:benchmarks}
\end{table}

\section{Human evaluation protocol}

\subsection{Measuring expressivity preservation}

We introduce a novel protocol to measure expressivity preservation in speech-to-speech translation systems. Human evaluations of synthetic speech typically rely on MOS measures of \textit{naturalness}, \textit{clarity of speech}, or  \textit{sound-quality}. Such measures provide little insight into the question of ``preservation" - whether or not characteristics of source- are shared (preserved) in target-speech. Moreover, while MOS measures may be related to expressivity, we believe they are generally too coarse-grained to be useful in an analysis of speech prosody and/or emotion.

\subsection{Protocol}
\label{ssec:protocol}
Our protocol requires bi-lingual annotators\footnote{\label{annotators}Annotators are recruited via a third-party vendor and must pass language fluency tests in order to be considered for the study.}  to rate the similarity of source- and target-audio across four ``core" and two ``auxiliary" expressivity aspects. Internal qualitative research identified the four aspects of \textit{emphasis},  \textit{intonation}, \textit{rhythm}, and  \textit{emotion} as primary focuses for expressivity preservation. Additionally, the protocol includes the two auxiliary aspects of \textit{manner} and \textit{meaning}\iftoggle{arxiv} {\footnote{A full justification of these aspects is beyond the scope of the current work, however we define and motivate them briefly sections \ref{core_aspects} and \ref{auxiliary_aspects}, including a short discussion of their limitations in \ref{framework_considerations}}}. We assess \textit{naturalness} in a separate study and it is not included as one of our expressivity aspects.

The first three aspects of \textit{emphasis}, \textit{intonation}, and \textit{rhythm} are oriented toward more  ``local" or ``prosodic" features of speech while \textit{emotion} is considered to be the most ``global" aspect. (Overall, their presentation here approximates a most-local to most-global ordering.) We provide summaries of each dimension in Table \ref{tab:protocol}.

\begin{table}[t]
	\centering
	\caption{Human protocol expressivity dimensions.}
	
	\centering
	\vspace*{-0.3cm}
	\begin{tabular}{ p{1cm} l p{5cm} }
		\toprule
		Aspect & Type &  Summary \\
		\midrule
		Emphasis & Core & Token-level acoustic variation signaling contrasting material. \\
		Intonation & Core & Utterance- or phrase-level rise and fall of the voice while speaking. \\
		Rhythm & Core & Patterns of speech-rate and pausing. \\
		Emotion & Core & Overall feeling or state of the speaker. \\
		Overall-Manner & Auxiliary & Core-aspects in aggregate. \\
		Meaning & Auxiliary & Semantics via lexical-content. \\
		\bottomrule
	\end{tabular}
	\vspace*{-0.5cm}
	\label{tab:protocol}
\end{table}
\iftoggle{arxiv} {
\subsubsection{Core aspects} \label{core_aspects}
The first three aspects of \textit{emphasis}, \textit{intonation}, and \textit{rhythm} are oriented toward ``local" or ``prosodic" features of speech. While we consider \textit{emphasis} to be a token-level property (characterizing a token's salience via elements such as volume or duration), \textit{intonation} and \textit{rhythm} are phrase- and utterance-level properties. In our protocol, \textit{intonation} describes the overall rise and fall of the voice while speaking, while \textit{rhythm} describes the speed, pacing and pauses in speech. We consider the fourth core aspect, \textit{emotion},  to be ``global" in nature as it characterizes speaker-state.

\subsubsection{Auxiliary aspects} \label{auxiliary_aspects}
In addition to our four core aspects, we introduce a question to assess \textit{meaning} (directed at translation quality via lexical-content) as well as \textit{manner}. In the current setting we include a question on \textit{meaning} for two reasons. First, as a filter in the case of semantically unaligned audio - we want to provide annotators with a mechanism to indicate ``semantic" discrepancies between source and target. Second, as a sanity check - we do not expect significant differences in ratings of \textit{meaning} between the baseline Vanilla TTS model or Proposed model. 

Our second auxiliary aspect, \textit{manner}, was considered as a possible composite measure of the core aspects. That is, to provide a single point of comparison which could account for variance located in the other four core aspects. In defining \textit{manner}, annotators are asked to consider the four core aspects in aggregate and provide a single rating.
}

\begin{table*}[t]
	\centering
	\caption{Main results. The Holistic Cascade system outperforms the Vanilla TTS baseline (higher ratings are better) on both datasets. This holds even with ground-truth (GT) input to the TTS model. * denotes statistical significance at $\alpha_{\text{adj}} < 0.0023$ (after Bonferroni correction).}
	
	\centering
	\vspace*{-0.3cm}
	\begin{tabular}{ c c c | c c c c | c c | c }
		\toprule
		\multirow{2}{*}[0pt]{Dataset} & \multirow{2}{*}[0pt]{Input text} & \multirow{2}{*}[0pt]{System} & \multicolumn{4}{c|}{Core aspects} & \multicolumn{2}{c|}{Auxiliary aspects} & \multirow{2}{*}[0pt]{Naturalness} \\
		& & & Emotion & Emphasis & Intonation & Rhythm & Manner & Meaning &  \\
		
		\midrule
		
		\multirow{5}{*}[0pt]{Heroes} & \multirow{2}{*}[0pt]{S2T} & Vanilla TTS & 2.379  & 2.901 & 2.462 & 2.297 & 2.302 & 3.769 & 3.050 \\
		& & Holistic Cascade & *3.188 & *3.419 & *3.075 & *3.231 & *3.027 & 3.785 & *3.515 \\
		\cmidrule(lr){2-10}
		& \multirow{3}{*}[0pt]{GT} & Vanilla TTS & 2.478 & 3.049 & 2.620 & 2.511 & 2.435 & 3.810  & 3.150 \\
		& & Holistic Cascade & *3.044 & 3.317 & *3.061 & *3.144 & *2.939 & 3.844 & 3.355 \\
		\cmidrule(lr){3-10}
		& & Human-reference & 3.883 & 3.888 & 3.856 & 3.872 & 3.830 & 3.867 & 4.525 \\
		
		\midrule
		
        \multirow{2}{*}[0pt]{\makecell{Mined\\audiobook}} & \multirow{2}{*}[0pt]{S2T} & Vanilla TTS & 3.031 & 3.296 & 2.959 & 2.714 & 2.776 & 3.867 & 3.170 \\
		& & Holistic Cascade & 3.108 & 3.307 & 2.960 & 2.989 & 2.886 & 3.852 & 2.910 \\
		\bottomrule
	\end{tabular}
	\label{tab:main-results}
\end{table*}
\begin{table}[t]
	\centering
	\caption{Mean ratings across all aspects increase as more global- and local- prosodic transfer components are included in the models. Current results examine the Heroes dataset with S2T input text.}
	
	\centering
	\vspace*{-0.3cm}
	\begin{tabular}{ c c | c c c c c }
		\toprule
		Global & Local & Emo. & Emp. & Int. & Rhy. & Man. \\
		\midrule
		\xmark & \xmark & 2.629 & 3.082 & 2.680 & 2.507 & 2.504 \\
		\xmark & \cmark & 2.864 & 3.185 & 2.717 & 2.978 & 2.690 \\
		\cmark & \xmark & 2.984 & 3.287 & 2.973 & 2.952 & 2.888 \\
		\cmark & \cmark & 3.114 & 3.348 & 3.032 & 3.121 & 2.951\\
		
		\bottomrule
	\end{tabular}
	\label{tab:results-local}
\end{table}

\iftoggle{arxiv} {
\subsubsection{Framework considerations} \label{framework_considerations}
Our current framework attempts to disentangle expressive features of speech along four core aspects (\textit{emphasis},  \textit{intonation}, \textit{rhythm}, and  \textit{emotion}). This approach has clear limitations. For example, it is not the case that expressive aspects necessarily occur independently of one another. However, asking an annotator to consider aspects in isolation (e.g. separate questions for \textit{emphasis} and \textit{intonation}) implicitly represents them this way. Likewise, asking an annotator to compare similarity along these aspects implies a direct mapping - if a particular word is emphasized in the source it should be emphasized in the target. However, language- or culture-specific characteristics (e.g. tonality, word order) may prevent such mappings from being valid (or naturalistic). The degree to which these issues and others may be present is a function of the particular language pair being studied. In the case of our current language pair, Spanish and English, we believe these issues are present, but surmountable, due to their similarities. By this we mean the overlap between their dominant-word-order and their non-tonality.

Despite these limitations, our protocol presents an important step toward establishing basic measures of expressivity preservation in the translation setting - moving beyond monolingual measures of \textit{naturalness}.
}

\subsection{Annotation}
Annotators first undergo calibration - training on a curated set of example pairs and ratings developed by the authors. After completing calibration, annotators may begin the study. For a given source-target pair, annotators are asked to first listen to the source, then target audio. If they find either source- or target-audio to be garbled or unclear they provide a single binary response and skip the remaining questions. For clear audio, annotators provide ratings using a 4pt likert scale ranging from 1-\textit{very different...}, 2-\textit{...some similarities, but more differences}, 3-\textit{...some differences, but more similarities}, to 4-\textit{very similar}. The order of audio pairs are randomized. However, aspect question ordering is fixed - annotators first rate the similarity in \textit{meaning}. If an audio pair receives a score of 1 for \textit{meaning}, annotators are asked to skip remaining questions. Following \textit{meaning}, annotators rate the similarity of \textit{emphasis}, \textit{intonation}, \textit{rhythm}, \textit{emotion}, and \textit{manner}. We do not expect an effect of aspect-ordering (this is not studied explicitly) and the current order approximately follows a design of more local to more global aspects.

\noindent{\textbf{Multi-grading and data preparation}}
In the current study, each source-target pair received five annotations (from separate annotators). Audio pairs in which the source- or target-audio was garbled or unclear ($<4$\% of samples), or pairs in which the majority of annotators indicated semantic-mismatch ($<8\%$ of samples following the audio-quality filter) were removed. Additionally, all annotations for one annotator were removed after the authors observed uniform ratings across systems. Such biased ratings indicated the annotator had not undergone sufficient calibration. The issue was not observed in any other annotators.

\noindent{\textbf{Score calculation}}
Item-level scores are computed by taking the median rating across the five annotations for a given sample pair. System-level scores are then computed by taking average of item-level scores.

\vspace*{-0.3cm}
\section{Experimental results}

In our experiments we focus on the Spanish-English direction. We compare the \textbf{Holistic Cascade} system to a \textbf{Vanilla TTS} baseline that does not use any prosody transfer techniques described in Sec.~\ref{sec:method}. To compare fairly, we include scenarios where the input text to the TTS model is (1) the true text from the reference English ground truth (\textbf{GT}) and (2) the translated text from the S2T model (\textbf{S2T}).

\subsection{Main results}
Table~\ref{tab:main-results} shows the main results. We performed a Wilcoxon Signed-Rank test to compare the Holistic Cascade and Vanilla TTS baseline models across all aspects with Bonferonni correction for multiple-tests. On the Heroes dataset, the Holistic Cascade model was rated higher than the Vanilla TTS baseline in all settings. Differences reached statistical significance in all but one setting (GT + \textit{emphasis}). Differences across systems were negligible for the \textit{meaning} aspect, as expected. Results for human-reference audio approached ceiling across all expressivity measures (4pt-scale) and exceeded both models in \textit{naturalness} ratings (5pt-scale).

\iftoggle{arxiv} {
\subsection{Parity on Mined Audiobooks data}
}
Notably, model differences are reduced (and do not reach significance) on the Mined Audiobooks data. Further investigation is needed to understand this particular result.
One factor is that the Mined Audiobook dataset, via internal inspection, appears less expressive than the Heroes dataset, thus there may not be room for the prosody transfer techniques to improve. Also, Heroes data and the Mined Audiobooks data differ in distributional characteristics. Of particular interest is the difference in segment length: as shown in Table~\ref{tab:benchmarks}, the median length for Mined Audiobooks target-audio is double that of the Heroes set. which may impact both modeling- and annotation-difficulty. 
% \al{Why do we not have human-reference results for mined audiobook in Table~\ref{tab:main-results}?}

\iftoggle{arxiv} {
Longer audio may impact results in multiple ways. Anecdotal evidence suggests that from an annotation perspective, as the length of audio increases, comparisons become increasingly difficult. While a full analysis should account for other sources of variation, we group audio segments by rounding their duration to the nearest integer and compute the Pearson's correlation between duration and average agreement for that duration grouping (taken as the proportion of samples in which a majority of annotators agreed upon a single rating). Under this informal analysis, results indicate a negative relationship $r(8) = -.61, p > .05$ between duration and agreement - as duration increases, agreement decreases. Future work should examine this relationship in more detail - both to account for other sources of variation, but also to establish annotation guidelines to ensure annotation quality remains high and the task manageable for annotators.
}

% \subsection{Impact of input text}
% A common concern for cascade systems is error accumulation. In our system, it is possible that the incorrectly translated text degrades the final performance. However, as in Table~\ref{tab:main-results}, our proposed method does not deteriorate with the S2T text. 

\subsection{Impact of local prosody transfer}

To verify the effectiveness of the local prosody transfer techniques, we compared the performance of systems with and without global and local prosody transfer. For pairwise combinations of \textit{global} and \textit{local} components we calculate aggregate system scores for that particular parameterization. Results indicate that including all the components leads to gains in expressivity preservation, shown in Table~\ref{tab:results-local}. 

% \subsection{Analysis on the metrics}

% \section{Conclusion and Future work}
\section{Conclusion}

In this work, we address existing limitations in expressive S2ST, presenting a holistic cascade system which incorporates multiple prosody transfer components, curate a benchmark test set, and propose a human evaluation protocol to assess expressivity preservation along multiple dimensions.
Results of our study indicate that expressivity preservation in S2ST systems can be assessed by bi-lingual annotators and that including multiple prosodic transfer components in a single system improves over more limited approaches\footnote{The Heroes and Mined Audiobook datasets will be made available publicly. Please check the link in footnote \ref{s2t}.}.

\iftoggle{arxiv} {
Although we tried to address previous limitations as much as possible, there is still much to improve.
First, the proposed local prosody transfer techniques rely on the accuracy of various modules, including the forced aligner, f0 extractor, and cross-lingual word aligner. Simpler designs should be considered to reduce the dependencies between modules.
Also, we only consider one language direction in our experiments, thus it is not clear to what extent our approach generalizes (either the system or the protocol). Future work should consider more language directions including pairs which may not have consistent word-order or use of tonality.
Lastly, the cascade nature of the proposed system hinders its application to translation directions that involve unwritten languages. It is therefore essential to develop toward direct expressive S2ST approaches.
% Lastly, we believe there may be an impact of audio duration - we should look more closely into this.
}

\section{Acknowledgement}
We thank Thilo Koehler for his help with the in-house forced aligners, Adam Polyak for helping with the Blizzard Challenge dataset, Ilia Kulikov and Yu-An Chung for the fruitful discussions, and Carleigh Wood for assistance in data collection.

\bibliographystyle{IEEEbib}
\bibliography{refs}

\iftoggle{arxiv} {
    \vfill\pagebreak
    \appendix
    % \onecolumn
    \section{Understanding the Heroes dataset with emotion labeling}
\label{sec:heroes_emotion_label}

To better understand the Heroes dataset, we conducted an emotion labeling study for Spanish and English audio. Annotators were asked to provide up to seven labels (from a pre-defined set derived from the emotion wheel \cite{moors2013appraisal}) for a single audio example. Note that labels in this study were gathered for Spanish or English audio in isolation, unlike our expressivity protocol which asks them to compare audio between the two languages. Each example received three annotations.

For a given example, we define a "top label" label (or labels) which received the most votes by all annotators. Table~\ref{tab:top_labels} shows the average number of top labels by language. On average, there are 1.8 top emotions in English and 1.6 in Spanish, respectively. This indicates that each sample receives more than one emotion label, suggesting the multi-dimensionality of speech emotion \cite{mixed-emotion, mixed-emotion-2}.

\begin{table}[ht]
	\centering
	\caption{Statistics of the number of top labels by language in the Heroes dataset.}
	
	\centering
	\begin{tabular}{ c | c c }
		\toprule
		Language & mean & std. \\
		\midrule
		English & 1.83 & 1.19 \\
		Spanish & 1.63 & 1.09 \\
		\bottomrule
	\end{tabular}
	\label{tab:top_labels}
\end{table}

In Figure~\ref{fig:emotion_label_distribution}, we show the distribution of top labels for each languages. First, we observe that the distributions of English and Spanish are different. In addition to the noises aroused by the dubbing process and the perceptual difference of the annotators, we believe that each language has its own prior distribution over the emotion spectrum. Also, we find that the "neutral" label was commonly selected, even accounting for the most selected emotion for the Spanish audio samples. To further investigate the emotion consistency in the Heroes dataset, we look into the number of sample pairs with the number of overlapped emotion labels between the English and Spanish sample pairs over certain amounts. Table~\ref{tab:overlapped_emotion_labels} shows the results. We find that only 233 out of the total 407 samples have at least 1 overlapped emotion label. 

\begin{figure}[ht]
	\centering
	\includegraphics[width=0.8\columnwidth]{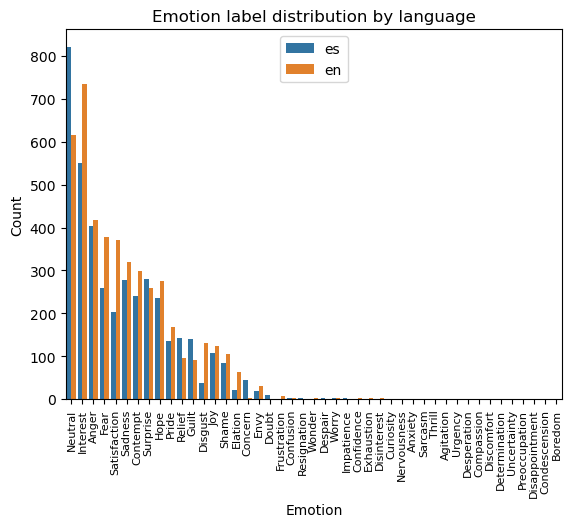} 
	\caption{The distribution of the emotion labels on the English and Spanish audio of the Heroes dataset. \label{fig:emotion_label_distribution}}
% 	\vspace{-0.3cm}
\end{figure}

\begin{table}[ht]
	\centering
	\caption{Cumulative number of samples with more than a certain amount of overlapped emotion labels between the English and Spanish sample pairs in the Heroes dataset.}
	
	\centering
	\begin{tabular}{ c | c }
		\toprule
		\# overlapped emotion labels & \# samples \\
		\midrule
		0 & 407 \\
		1 & 233 \\
		2 & 23 \\
		3 & 1 \\
		4 & 0 \\
		\bottomrule
	\end{tabular}
	\label{tab:overlapped_emotion_labels}
\end{table}

Although the above mentioned two observations may seemingly indicate that the Heroes dataset is lowly emotional and lowly emotional consistent, we would like to emphasize that emotion only accounts for one of the four core aspects we defined in Sec.~\ref{ssec:protocol}. As shown in Table~\ref{tab:main-results}, the Heroes dataset is still considered consistent in terms of the four expressivity aspects we defined.

\section{Full Human Evaluation Protocol}

Below, we provide the complete expressivity human evaluation protocol used in our Spanish/English study, including annotator instructions and likert questions for each of the aspects. Note that we have tried to reproduce orthographic markers such as bolding and italicization. For clarity, all likert response options are single-choice (an annotator may only select one item from i. - iv.).

\subsection{Overview}
In this task, you will listen to pairs of audio segments. Each pair will consist of one Spanish segment and one English segment. You will then rate how similar the two segments sound in terms of prosody (defined as patterns of emphasis, intonation, and rhythm) and emotion present in the speech. Some of these segments may be direct translations of one another and some may not.

\subsection{Process}
\begin{enumerate}
    \item Listen to the Spanish segment from start to finish. Then listen to the English segment from start to finish. 
    \begin{itemize}
        \item If either of the segments is very garbled or unclear, please check the box “audio issues” and skip the item.
        \item If the segments have the same or similar content, but one of them has more words at the end or beginning, please only consider the content shared between the two segments. If the difference in the amount of content is greater than a few words, please skip to the question on Semantics and select a score of 1 (Very different). Where this occurs, you will not be answering questions related to the other four expressivity dimensions.
        \item Pay attention to the \textbf{emphasis}, \textbf{intonation}, and \textbf{rhythm} of both segments, as well as the \textbf{emotions} being conveyed.
        \item Two segments can be similar in the presence or absence of the aspects of interest. That is, if two sentences are both equally neutral in any of the categories, we can also consider them to be “similar.” For example, we would consider two segments as being similar in emphasis if they both emphasize the same word or if no single word is emphasized in either segment.
    \end{itemize}
    \item Please answer the following questions:
    \begin{enumerate}
        % Semantics
        \item The \textbf{semantics} of an utterance refers to the literal meaning of the words disregarding the manner in which they are spoken. For example, the sentence “There is a green apple” in English has a different meaning from “Hay una manzana roja” in Spanish.  Thinking only about the words used in each segment and the actions, objects or concepts they refer to - \textit{do each of these segments have similar semantics?}
        \begin{enumerate}
            \item The two segments are \textbf{very different} in their meaning - they refer to different objects, actions or concepts and the relationships between them.
            \item The two segments have some similarities, but \textbf{more differences} in their meaning.
            \item The two segments have some differences, but \textbf{more similarities} in their meaning.
            \item The two segments are \textbf{very similar} in their meaning - they could be paraphrases of one another.
        \end{enumerate}
        % Emphasizing
        \item \textbf{Emphasizing} a word is similar to "bolding it" in written text, calling attention to the word. For example, placing emphasis on “You lied to me” in English and “Tu me mentiste” in Spanish by increasing the word's volume or length might indicate that the speaker is surprised that their interlocutor lied to them, but that it might not have been expected from others. \textit{Do the two segments place emphasis on the same or similar words and concepts?}
        \begin{enumerate}
            \item The two segments sound \textbf{very different} in their emphasis - basically none of the emphasized aspects are shared.
            \item The two segments share some similarities, but \textbf{more differences}.
            \item The two segments have some differences, but \textbf{more similarities}.
            \item The two segments sound \textbf{very similar} in their emphasis - basically all of the emphasized aspects are shared.
        \end{enumerate}
        % Intonation
        \item \textbf{Intonation} characterizes the rise and fall of the voice while speaking. For example, “You lied to me?” in English and “¿Tu me mentiste?” in Spanish both having a sharp rising tone may indicate a shocked question. \textit{Do the two segments sound similar in terms of intonation?}
        \begin{enumerate}
            \item The two segments sound \textbf{very different} in their intonation - basically none of the intonation characteristics are shared.
            \item The two segments share some similarities, but \textbf{more differences}.
            \item The two segments have some differences, but \textbf{more similarities}.
            \item The two segments sound \textbf{very similar} in their intonation - basically all of the intonation characteristics are shared.
        \end{enumerate}
        % Rhythm
        \item \textbf{The rhythm} of an utterance describes its speed, pacing (i.e. changes in speed), and pauses. A speaker pausing or elongating/shortening words can impact rhythm. For example “You -- lied to me” having a pause after “you”, and “¿Tu -- me mentiste?” having a pause after “tu” are distinct from "You lied to -- me?". A speaker speaking quickly or slowly throughout the sentence, or speeding up/slowing down at certain parts of the sentence, also impacts rhythm. \textit{Do the two segments sound similar in terms of rhythm?}
        \begin{enumerate}
            \item The two segments sound \textbf{very different} in their rhythm - basically none of the rhythmic aspects are shared.
            \item The two segments share some similarities, but \textbf{more differences}.
            \item The two segments have some differences, but \textbf{more similarities}.
            \item The two segments sound \textbf{very similar} in their rhythm - basically all of the rhythmic aspects are shared.
        \end{enumerate}
        % Emotion
        \item \textbf{Emotion} characterizes the overall feeling of the speaker.  For example, a speaker may sound angry, shocked or happy (to name just a few emotions) while speaking. \textit{Do the two segments sound similar in the emotions they convey?}
        \begin{enumerate}
            \item The two segments sound \textbf{very different} in the emotions conveyed - basically none of the emotion aspects are shared.
            \item The two segments have some similarities, but \textbf{more differences}.
            \item The two segments have some differences, but \textbf{more similarities}.
            \item The two segments sound \textbf{very similar} in the emotions conveyed - basically all of the emotion aspects are shared.
        \end{enumerate}
        \item Considering the \textbf{overall manner} in which these two segments are spoken, that is, the emphasis, intonation, rhythm and emotion taken together - \textit{how similar are they?}
        \begin{enumerate}
            \item The two segments sound \textbf{very different} in their overall manner. 
            \item The two segments have some similarities, but \textbf{more differences}.
            \item The two segments have some differences, but \textbf{more similarities}.
            \item The two segments sound \textbf{very similar} in their overall manner.
        \end{enumerate}
    \end{enumerate}
\end{enumerate}

}

\end{document}